\documentclass[letterpaper, 10 pt, conference]{ieeeconf}  %

\IEEEoverridecommandlockouts                              %

\usepackage{graphics} %
\usepackage{epsfig} %
\usepackage{mathptmx} %
\usepackage{times} %
\usepackage{amsmath} %
\usepackage{amssymb}  %
\usepackage{algorithm, algorithmic}
\usepackage{hyperref}

\usepackage{amsthm}
\usepackage[T1]{fontenc}
\usepackage{subcaption}
\usepackage{xcolor}
\usepackage{url}
\usepackage{hyperref}

\usepackage{array} %
\usepackage{multirow} %
\usepackage{graphicx} %
\usepackage{tabularx} %

\title{\LARGE \bf
DECAF: a Discrete-Event based Collaborative Human-Robot Framework for Furniture Assembly
}

\author{ Giulio Giacomuzzo$^{1}$, Matteo Terreran$^{1}$, Siddarth Jain$^{2}$, Diego Romeres$^{2}$ \thanks{$^{1}$Department of Information Engineering, Università di Padova, Italy \newline {\tt\small \{giacomuzzo, matteo.terrean\}@dei.unipd.it}\newline$^{2}$MERL \newline {\tt\small \{sjain ,romeres\}@mer$^{2}$ l.com}}}

\begin{document}

\twocolumn[{%
    \renewcommand\twocolumn[1][]{#1}%
    \maketitle
    \begin{center}
        \centering
        \vspace{-5mm}
        \includegraphics[width=1.0\textwidth]{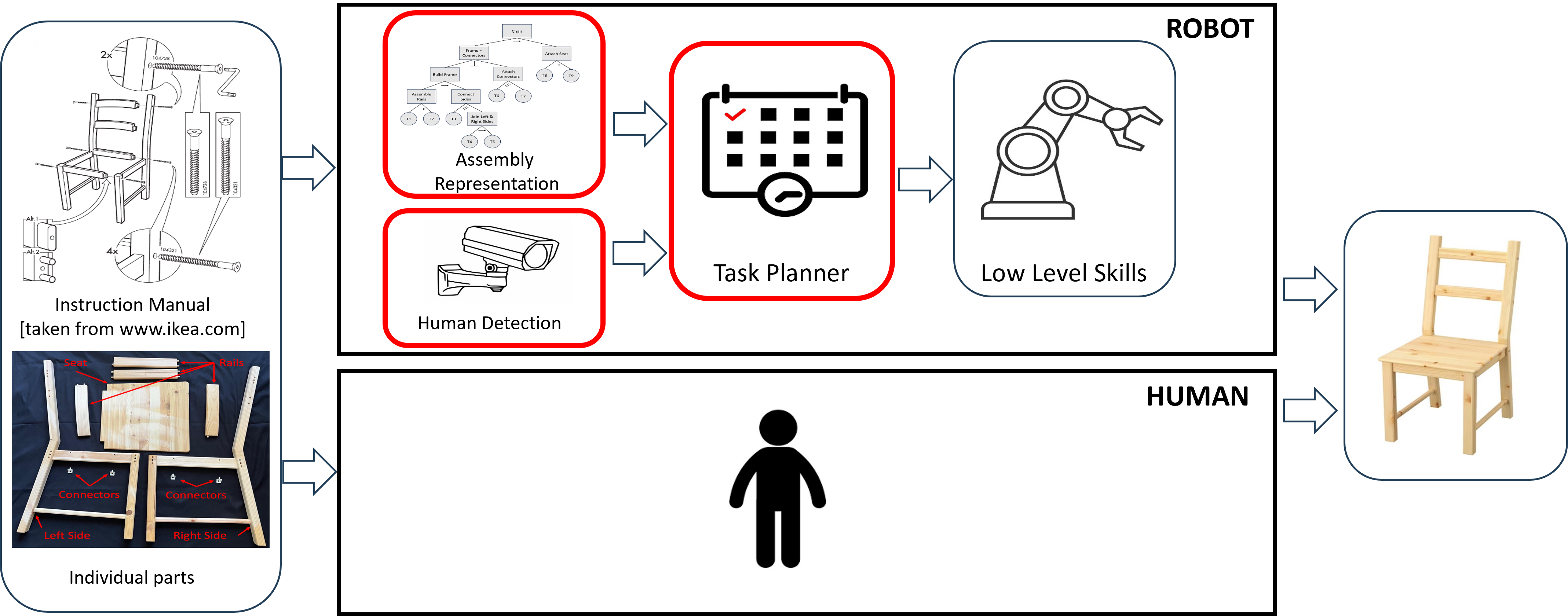}
        \captionof{figure}{Schematic of a complete flow to achieve collaborative furniture assembly. This work focuses on the components highligted in red.}        
        \label{fig:intro_fig}
    \end{center}%
}]
\footnotetext[1]{Department of Information Engineering, Università di Padova, Italy \newline {\tt\small \{giacomuzzo, matteo.terrean\}@dei.unipd.it}}
\footnotetext[2]{Mitsubishi Electric Research Laboratories (MERL), Cambridge, MA, USA. {\tt\small \{sjain ,romeres\}@merl.com}}

\begin{abstract}
This paper proposes a task planning framework for collaborative Human-Robot scenarios, specifically focused on assembling complex systems such as furniture. The human is characterized as an uncontrollable agent, implying for example that the agent is not bound by a pre-established sequence of actions and instead acts according to its own preferences. Meanwhile, the task planner computes reactively the optimal actions for the collaborative robot to efficiently complete the entire assembly task in the least time possible.
We formalize the problem as a Discrete Event Markov Decision Problem (DE-MDP), a comprehensive framework that incorporates a variety of asynchronous behaviors, human change of mind and failure recovery as stochastic events. Although the problem could theoretically be addressed by constructing a graph of all possible actions, such an approach would be constrained by computational limitations. The proposed formulation offers an alternative solution utilizing Reinforcement Learning to derive an optimal policy for the robot. Experiments where conducted both in simulation and on a real system with human subjects assembling a chair in collaboration with a 7-DoF manipulator.
\end{abstract}

\section{Introduction}\label{sec:introduction}

In recent years, the area of human-robot collaboration has gained increasing attention because of the advancement of robotics and artificial intelligence. The synergistic integration of human dexterity and robotic precision holds great potential, especially for small- and medium-sized enterprises with high-mix and low-volume productions. In this context, the assembly of furniture represents a challenging yet promising test-bed application, since it involves long-term action sequences, dexterous manipulations, precise alignments, transportation of cumbersome objects and three-hands operations. Reactive planners for the robot actions play a crucial role in promoting smooth collaboration, yet robust frameworks for long-term tasks in unpredictable environments remain challenging and warrant further investigation. For effective collaboration, robots must adapt to human behavior while maintaining coordination, although explicit coordination may slow interaction. Additionally, unexpected events like task failures or sudden human change of mind can hinder team effectiveness and safety, and thus need to be considered and handled at planning time. %

Several solutions have recently been explored in the literature to address the aforementioned challenges. If the assembly task can be decomposed into fully independent sub-tasks, they can be assigned offline either to the human or to the robot, which then work separately of each other, sharing only the workspace \cite{fusaro2021human}. This situation, however, is unrealistic, as the interdependence between actions, which is always present in real world assembly scenarios, requires the agents to be mutually adaptive and take active roles in the action selection.

To address the coordination between agents, a large part of the literature treat the long-term collaborative assembly planning as a multi-agent scheduling task \cite{dhanaraj2023contingency, flexHRC, johannsmeier2016hierarchical}. Within this solutions, the human and the robot are both considered controllable agents, and the assembly sub-tasks are assigned according to some optimality criterion, in order to reach the shared goal. Despite being effective, these approaches typically require the human agent to follow a prescribed plan or behave according to an optimal policy, which could limit both the efficiency and the naturality of the collaboration. 

Single-agent frameworks have been also proposed, mainly relying 
on a leader-follower paradigm, where the robot actions are passively determined by the human leader through a dedicated interface \cite{fukui2009development}, or learned from demonstrations \cite{zhu2018robot}. Few works have overcome the leader-follower paradigm, enabling the robot to optimally choose its actions and adapt to the human behaviour \cite{gottardi2023dynamic, ramachandruni2023uhtp, cheng2020towards,cheng2021human}. These methods, generally, lack robustness against uncertainties like task failures and human changes of mind, common in practical applications. Additionally, they seldom account for agents collaborating on the same sub-task.

In this paper, we present a novel single-agent framework for planning long-term collaborative assembly tasks, such as furniture assembly, called Discrete-Event based Collaborative Human-Robot Assembly Framework for Furniture (DECAF). In the proposed framework the robot agent observes the human, performs both joint and independent actions maximizing the task efficiency while reacting to unpredictable events. In details, we rely on a Hierarchical Task Model (HTM) to encode the interdependence between the assembly sub-tasks, and on a Bayesian Inference module to observe the human agent behaviour. The collaborative assembly is then modelled as a Discrete-Event Markov Decision Process (DE-MDP), which allows to easily handle actions of different duration, as well as uncertain contingencies such as failures and changes of mind. To solve the DE-MDP and obtain a planning policy, we propose both a deterministic decision graph and a Reinforcement Learning (RL) method. 

\subsection{Contribution}
Our contribution can be summarized as follows: i) We propose a novel DE-MDP formulation to model long-term collaborative assembly tasks, which accounts for asynchronous joint and independent actions with uncertain duration, and unpredictable events such as task failures and changes of mind.
ii) We present DECAF, a complete planning framework which comprises the DE-MDP model, a HTM description of the assembly and a Bayesian Inference module to discover human intention. iii) We assess our framework in both simulated and real-world environments. DECAF is validated through a real chair assembly task involving 10 human subjects. Results show decreased completion times compared to a human solving the task alone, along with reduced physical effort and an enhanced assembly experience for the human operator.

\section{Related Works}
There exists an extensive literature on planning, scheduling and task assignment for human-robot assembly task. Several approaches model the assembly task by means of AND/OR graphs and solve the planning as a search problem \cite{de1990and, flexHRC, johannsmeier2016hierarchical, knepper2014distributed}. Similarly, Behavior Trees have been also considered as task representation \cite{fusaro2021integrated, fusaro2021human, iovino2022survey}. Despite they proved to be effective, all these works require the human to follow the specific plan prescribed by the scheduler. Moreover, all these methods are not robust to uncertain and unpredictable events.

Robustness to task failures is considered in \cite{dhanaraj2023contingency}, where the task is represented by means of an HTM and the cooperation is modeled as a multi-agent concurrent MDP. The MDP is solved by building a decision graph which also considers sub-task failure probability, and the optimal path is selected defining a Mixed Integer Linear programming Optimization. However, this framework requires the human to adhere to pre-computed plans and does not consider joint actions. Additionally, integrating contingencies different to task failures proves challenging, as they should be manually added to the graph construction.

A different approach is taken in \cite{you2023robust}, where a robust planning algorithm based on Partially Observable MDP is developed. The authors concentrate on deriving a realistic human policy, considering the interplay between the decision-making processes of both agents, which directly affects human intention. Their solution enables both agents to execute individual and joint actions. However, they overlook actions of varying durations and robustness to uncertainty.

More similarly to our work, single-agent formulations have been explored in \cite{cheng2020towards}, \cite{cheng2021human} and \cite{ramachandruni2023uhtp}. 
In \cite{cheng2020towards} a Robust Plan Recognition and Trajectory Prediction
is proposed based on an HTM representation, where the human's intentions and plan are recognized using a perception module. The robot recognizes the human plan and performs its actions accordingly, but its role is limited to assistive operations, without the possibility actively contribute to the task advancement.

In \cite{cheng2021human} the same authors propose a collaborative framework where the HTM task scheduling is optimized to minimize the total task completion time and promote workspace separation. However, within this framework the possibility to collaborate together on the same joint task is not considered and the task is optimized only over a limited time horizon.

Similarly in \cite{ramachandruni2023uhtp}, the authors rely on a modified HTM task representation and their goal is to provide an adaptive framework where the overall assembly time is minimized. The robot observes the human choice and selects its action in order to minimize the cost of the entire framework, averaged over all the possible human actions. However, their solution relies on a graph based strategy, which is not suitable to handle failures and other stochastic events. 

\section{Problem Formulation}\label{sec:problem_formulation}
In this section, we formalize the long-term collaborative assembly planning problem we aim to solve, and we highlight the requirements and assumptions of our framework.
Assembly tasks are defined as a hierarchical set of atomic tasks, hereafter denoted as actions, which must be executed complying with a set of ordering constraints. We assume that two agents, a human and a robot, cooperate to solve the task. Furthermore, we assume actions to be either individual or joint. Individual actions can be executed by a single agent, be it the human or the robot, while joint actions necessitate simultaneous execution by both the agents. The framework is robust to actions of different duration, contingent upon the specific action and the agent undertaking it. 

Despite the aforementioned scenario includes two agents, we assume the human to be inherently uncontrollable. The robot can only observe the human actions and plan its own actions based on the state of the environment which comprises the assembly state and behavior of the human partner. Consequently, the problem is formulated as a single agent planning task. 

The setting depicted above introduces interesting challenges, which we aim to address within our framework.\\
\textbf{Synchronization:} the human and the robot can work in parallel on different actions but have also to perform joint actions together. Since actions have different duration, agent synchronization is required to perform joint actions.\\
\textbf{Human detection:} as the human behave uncontrolled and the robot has to adapt its actions to the human choices, human action detection is required. This inevitably introduces a delay into the robot action scheduling, which needs to be considered by the planner.\\
\textbf{Human change of mind:} the human could begin a task, and then suddenly change its mind and perform a new task. In that case, the robot should reschedule its plan. \\
\textbf{Failures:} The actions performed by the agents could fail, which requires the introduction of recovery strategies.

To address the aforementioned challenges, we consider the following assumptions. i) Joint actions can be chosen and initiated only by the human operator. In case the human chooses a joint action, the robot joins them as soon as it finishes the ongoing action. ii) We introduce an additional action called \textit{idle}, which can be performed by both the agents. When \textit{idle}, the agent is waiting, without contributing to the task advancement. We use the \textit{idle} as a synchronization mechanism before a joint action: when the human chooses a joint action, they remain idle until the robot has completed its action. When the robot finishes, the two agents perform the joint action together. iii) If the human action is not detected, the robot must remain \textit{idle}. v) Failures are detected at the end of the performed action, which includes actions that fail either before reaching the goal or due to incorrect execution. We do not account for failures occurring after an action has been successfully completed. %

\section{DECAF framework}\label{sec:method}

\subsection{Hierarchical Task Model for Assembly task description}\label{sec:HTM}
The DECAF framework relies on sequential/parallel Hierarchical Task Model (HTM) for task representation \cite{cheng2020towards}.
 A HTM is a tree structure where  the root node represents the entire assembly, while all the other nodes represent assembly subtasks. The leafs represent the actions to be executed. Each node can be categorized as parallel ($\parallel$), sequential ($\rightarrow$) or independent ($\perp$). Children of parallel nodes can be executed at the same time by the two agents, in any order. Children of sequential nodes must be individually executed in the specified order (from left to right). Finally, children of independent nodes must be individually executed but can be performed in any order. An example of HTM is in Fig.~\ref{fig:HTM_real}.

In our framework, the HTM is used to describe the sequential constraints on action executions, which are encoded in the DE-MDP described in Section \ref{sec:de_mdp}.

\subsection{Bayesian Inference for Human Intent Recognition}\label{sec:human_detection}
To recognize the human actions, DECAF relies on a Bayesian Inference module. We assume each action to be related to a specific object in the assembly scene. Then, we use Bayesian Inference to detect the object of interest for the human, which in the following will be denoted as \textit{goal}. 
The robot, during the assembly, task aims to infer the most likely goal $g^*$ for the human from the set of possible goals $\boldsymbol{g}$, given a set of observations. Following~\cite{jain2019probabilistic, jain2018recursive}, we formulate the intent inference problem as Bayesian filtering in a Markov model, which allows us to model the uncertainty over the candidate goals of the human agent as a probability distribution over the goals. 
We integrated human hand pose tracking through a deep learning model~\cite{zhang2020} finely-tuned within the experimental setup. %
We represent the goal $g_t$ as the query variable and the observed features $\Theta_0$,...,$\Theta_{t}$ as the evidence variables, where $\Theta_t$ is a $k$-dimensional vector of $k$ observations $\theta_t^i$, $i=1:k$, at time $t$. The uncertainty over goals is then represented as the probability of each goal hypothesis. In particular, we compute proximity and alignment to a goal as likelihood features, given the human hand pose and goal poses. We assume the conditional independence of observations, and thus, at time $t$, the belief $b_t(g)$ becomes, 
\begin{equation} 
\begin{split}
b_t(g) = P(g_{t} \mid \Theta_{0:t}) & \propto \prod_{\theta_t \in \Theta_t}^{}  P(\theta_t \mid g_t) \\  
& \sum_{g_{t-1} \in \boldsymbol{g} }^{}  P(g_t \mid g_{t-1}) b_{t-1}(g_{t-1}).
\end{split}
\end{equation}

The posterior distribution at time $t$, denoted $b_t$, represents the belief after taking the observations into account. The set of prior probabilities $P(g_{t=0})$, $\forall$$g$ $\in$ $\boldsymbol{g}$, initially represents the robot's belief over the goals. The beliefs then are continuously updated, by computing the posteriors, as more observations become available. Furthermore, $P(g_t \mid g_{t-1})$ is the conditional transition distribution of changing to goal $g_t$ at time $t$ given that the goal was $g_{t-1}$ at time $t-1$.  Finally, to predict the most likely goal $g^*_{t}$ $\in$ $\boldsymbol{g}$, we select the goal class that is most probable according to the maximum \textit{a posteriori} decision, 
\begin{equation}
\ g^*_{t} =  \arg\max_{g_t \in \boldsymbol{g}} P(g_t \mid \Theta_t).
\end{equation}

\subsection{Discrete-Event MDP}\label{sec:de_mdp}

In this section, we describe the proposed method to model and solve the collaborative assembly scheduling. Our method describe the system as a Discrete-Event Markov Decision Process (DE-MDP). A DE-MDP is an MDP in which the state transitions do not happen on a fixed time basis, but are determined by the occurrence of relevant events. In the case of collaborative assembly, the task advancement is related to the completion of actions which have different, possibly stochastic, durations. Moreover, the cooperation state is also modified by events such as the human action recognition or the human change of mind. 

Consider an assembly task composed by $N$ actions. The collaborative scheduling problem is modeled as a DE-MDP ($S$, $A$, $\mathcal{E}$, $\Gamma$, $\ell$, $T$, $R$, $\gamma$), where $S$ is the state space of the system; $A$ is the action space; $\mathcal{E}$ represents the set of possible events; $\Gamma(s)$ is the set of feasible events at state $s \in S$; $\ell(s,a,e)$ is the event lifespan function, which gives the probability distribution over the time after which the event $e$ is likely to occur at the current state $s \in S$ and robot action $a \in A$; $T(s, a, s', e)$ is the transition probability function which gives the probability of transitioning to state $s'$ given the current state $s$ the current robot action $a$ and the event $e$; $R(s, a, e)$ is the reward function and $\gamma$ the discount factor. 

Differing from the standard MDP framework where the state update occurs at every time step, the DE-MDP state transition happens only in the presence of an event. We use the index variable $k$ to represent the time instant of the state evolution. Moreover, $\Delta t ^ k$ denotes the time spent by the system to transition from state $s^k$ to $s^{k+1}$ and, for convenience, it is multiple of a discrete time step $\bar{T}$. Therefore, the variable time step, $\Delta t ^ k$, at which the DE-MDP state updates depends on the event lifespan $\ell(s^k, a^k, e)$, with $e$ being the event causing the system transition from $s^k$ to $s^{k+1}$. In the following, for notation simplicity, we will explicit the dependence on the index $k$ only when necessary.

In the following the elements of the DE-MDP are detailed. 

\paragraph{Actions} the actions set $A$ includes all the actions that can be executed by the robot. Let $\bar{A}=\{a_1,\dots,a_N\}$ be the set of all possible actions to complete the task. Let us also introduce the set $A_r=\{a_{N+1},\dots,a_{2N}\}$ be the set of recovery actions. These actions are not necessary to complete the task, but must be executed to recover from a failure. Finally, let us denote with $a_{idle}$ the \textit{idle} action. We model each action as a tuple $a_j=(o_j, \delta_j, p^f_j)$, where $o_j \in \{0,1,2,3\}$ denotes if the action can be performed by the human (0), by the robot (1), by both the agents (2) or is a joint action (3); $\delta_j = [\delta_j^h, \delta_j^r]$ represents the action duration when performed by the human ($\delta_j^h$) or by the robot ($\delta_j^r$). Regarding the action duration modeling, we assume $\delta_j^h$ and $\delta_j^h$ to be normally distributed around a nominal action duration value; $p^f_j \in [0,1]$ denotes the action failure probability. Note that in the case of joint actions, $\delta_j^h$ and $\delta_j^r$ are equal, while we assume the \textit{idle} action to last until the next triggering event happens. 
With this formalism, we define the action space as  $A=\{a_j \in \bar{A}| o_j \neq 0\} \cup \{a_j \in A_r| o_j \neq 0\} \cup \{a_{idle}\}$, which has cardinality $N_r$.
For future convenience, let us define also the set of human actions $A_h=\{a_j \in \bar{A}| o_j \neq 1\} \cup \{a_j \in A_r| o_j \neq 1\ \cup \{a_{idle}\}$, which has cardinality $N_h$.\\
\textbf{Feasible Actions:} $A_f$ is the set of all feasible actions for the two agents at the current state. The nonfeasible actions are the actions already completed, the actions for which the requirements are still not satisfied, the recovery actions for which the corresponding action is not failed, the current action being executed by the other agent, and the \textit{idle} action for the robot if the human is already \textit{idle}, to avoid infinite cycles. 

\paragraph{States} the system state is $s=(s_a, \,a^h, \,t^h, \,t^r, \,d)$ where $s_a \in \{-1,0,1\}^N$ is the actions execution indicator, the $i$-th component representing the $i$-th action being completed (1), not attempted yet (0) or failed (-1); $a^h \in A_h \cup \{unknown\}$ is the current human action, which is part of the state because the agent cannot be controlled. Note that the human action is considered \textit{unknown} only in case it has not been detected yet; $t^h \in \mathbb{N}$ is the time elapsed since the current human action started; $t^r \in \mathbb{N}$ is the time since the current robot action started; $d \in \{0, 1\}$ indicates if the current human action has been detected or not. 

\paragraph{Events} the set $\mathcal{E} = \{H, \,R, \,D, \,C\}$ contains the events triggering the state transitions, where $H$ represents the end of a human action, $R$ represents the end of a robot action, $D$ represents the human action detection and $C$ represents a human change of mind. 
As a modeling choice, we assume the probability of an event happening $p(e|s,a)$ can be fully determined by the current state $s \in S$ and robot action $a \in A$. In particular, if $e=C$, we denote with $p_c \in [0, 1]$ the probability of a human change of mind to happen. Then, $p(C|s,a)$ represents the probability of having a change of mind before the next $H$ or $R$ event. The probability of having a change of mind before the next $H$ event is $p_c$, as, if the change of mind happens, it will for sure happen before the end of the current human action. The probability of having a change of mind before the next $R$ event, instead, depends on the time at which the next $C$ and $R$ will happen. Let $\Delta_C$ and $\Delta_R$ be the time after which the next $C$ and $R$ events will happen, respectively. Note that $\Delta_C$ and $\Delta_R$ are random variables distributed as $\ell(s,a,C)$ and $\ell(s,a,R)$, respectively. Then, $C$ will happen before $R$ with probability $p_c p_{cr}$, with $p_{cr}$ being the probability $p(\Delta_C < \Delta_R)$. See the paragraph regarding the lifespan function for the description on how the lifespan distributions are defined. Note that $p_{cr} = 1$ if $\Delta_R > \Delta_H$. We can in general conclude that $p(C|s,a) = p_c p_{cr}$. Then, if $e=H$, the probability $p(H|s,a)$ is 0 if the robot finishes its action before the human i.e., $\delta^r_a - t^r$ < $\delta^h_{a^h} - t^h$, while it is $1-p(C|s,a)$ if the human finishes before the robot. Analogous reasoning applies for the event $e=R$. Finally, if $e=D$, $p(D|s,a) = 1$  if $d=0$, otherwise $p(D|s,a) = 0$.

\paragraph{Feasible event set} $\Gamma(s):S \rightarrow E \subseteq \mathcal{E}$ provides the set of feasible events given the current state $s\in S$. In particular we have: $\Gamma(s| d=0) = \{D\}$, namely if the human action has not been detected, the only feasible next event is the detection, while if the human action has been detected all the other events are feasible, namely $\Gamma(s| d=1) = \{H\,,R\,,C\}$. 

\paragraph{Event lifespan} the event lifespan function $\ell(s,a,e)$ provides a probability distribution over the time spent by the system in the current state $s \in S$ with the current robot action $a \in A$, before transitioning to the next state due to the event $e\in \mathcal{E}$, namely $\ell(s, a, e) = p(\Delta_e| s, a)$, with $\Delta_e$ being the time after which the event $e$ will occur given $s$ and $a$. 
In the assembly scenario we are considering, if $e = D$ then $\Delta_D = \bar{\Delta}_D$ with probability $1$, where $\bar{\Delta}_D$ represents the human action detection time, which we assume to be deterministic and known. 
If $e = H$, $ \Delta_H = \delta_{a^h}^h-t^h$. 
If $e = R$, $ \Delta_R = \delta_{a_r}^r-t^r$, 
where $\delta_{a^h}$ and $\delta_{a^r}$ represent the human's and robot's actions duration, while $t^h$ and $t^r$ represent the time elapsed from the human and the robot started their current action.
Note that, as $\delta_{a_h}$ and $\delta_{a_r}$ are random variables, also $\Delta_H$ and $\Delta_R$ are random variables with the same distribution as $\delta_{a_h}$ and $\delta_{a_r}$, with mean shifted by the elapsed times $t^h$ and $t^r$, respectively.  
Finally, we model the lifespan $\Delta_C$ of the event $e=C$ as a discretized truncated exponential distribution with support spanning the time interval from the detection time to the end of the human action. 

\paragraph{Transition} The transition function $T(s,a,s',e)$ provides the probability of transitioning to the next state $s' \in S$ given the current state $s$ and the current robot action $a$ due to the occurrence of the event $e$. We define in the following the transition probability for each possible event in the collaborative assembly scenario we are considering. If $e=D$, namely the event is a human action detection, the only effect on the state is the deterministic transition of $d$ from $0$ to $1$. If $e=H$, namely the event is a human action end, the transition will affect $s_a$, $a^h$, $d$ and $t^h$. In particular, $s_a$ will transition to $s_a' = s_a \pm 1_{a^h}$, with $+$ if the action ends successfully and $-$ if it fails, where $1_{a_j} \in \{0,1\}^{N+1}$ is the vector containing all zeros except for the component corresponding to $a_j$. $a^h$ will transit to the \textit{unknown} state, while $d' = 0$ and ${t^h}' = 0$. If $e=R$, the transition will affect $s_a$ and $t^r$, in particular $s_a' = s_a \pm 1_a$ and ${t^r}' = 0$. Finally, a change of mind $e=C$ will cause the human and consequently the robot to change action, namely ${a^h}'$ will transit to the \textit{unknown} state, $d$ will transit to $0$, while ${t^h}' = 0$ and ${t^r}'$ = 0.

\paragraph{Reward} as we aim to minimize the execution time, we model the reward as the negative transition time, namely $R(s^k, a^k, e)=-\Delta_t^k$. Different rewards could be possible, which for example consider human ergonomics or previous action correlation.

\subsection{DE-MDP Solutions}
Solutions to the DE-MDP provide a robot policy $\pi_\theta(\cdot)$ mapping the current state $s$ to the next robot action $a$. In this work we consider two solutions, one based on a deterministic decision graph and the other one obtained using RL.
\paragraph{Decision Graph} 
given the DE-MDP description, a decision graph can be built assuming a deterministic setting, namely considering the nominal values for actions duration and disregarding stochastic events such as the human change of mind. A decision graph is a directed graph in which each node represents a state, while each edge represent a robot action. It can be built starting from the initial state and computing, for each state, all the possible transitions, until the final state is reached. Each edge of the graph is weighted with the cost of the transition (e.g. the transition time). 
Then, any graph search algorithm (e.g. Dijkstra \cite{dijkstra1959note}) can compute, for each state, the minimum cost path to the final node. The union of all the paths is the optimal policy and it can be computed offline and stored in a look-up table. %
Nonetheless, the generation of the decision graph can be particularly inefficient or even infeasible both in terms of computational time and storage requirements, when the number of actions and possible transition increase. Moreover, stochasticity is not taken into consideration and the policy can only react to stochastic events.   

\paragraph{Reinforcement Learning}
Given the problem formalization as an MDP, RL algorithms are well suited to overcome the limitation of the approach based on the decision graph. In particular, in presence of predictable stochastic events, RL can learn the stochastic dynamics and take actions that consider also the possible future evolution. Moreover, parametric function approximators can be exploited in the presence of high dimensional state-action spaces to reduce the computational and storage burden.

\section{Simulated Experiments}\label{sec:results}
In this section we evaluate DECAF's advantages on simulated collaborative assembly scenarios. We designed HTMs representing both toy examples of increasing complexity and a real chair. First, we consider a deterministic setting, where no uncertainties are present. In such a scenario the graph based policy can be easily computed and represent the optimal solution. We show that, the policy learned with RL converges to the same optimal performance, and that both the DECAF policies outperforms greedy and random baselines.
Second, we introduce stochastic events such as failures and changes of mind, and we show that under these conditions the RL policy still provides a feasible solution, while the decision graph cannot be generated.
In both the aforementioned scenarios, we perform a Monte Carlo experiment composed of multiple trials, considering variable action duration and random human behaviour.  

In both settings, the RL policy is trained with the Proximal Policy Optimization (PPO) algorithm \cite{schulman2017proximal}. To this aim, we implemented the DE-MDP described in Section~\ref{sec:de_mdp} as a custom Gym environment \cite{towers_gymnasium_2023} and used the Stable Baselines library \cite{stable-baselines3} for policy training. The training process is sped up using action masking to limit the action exploration only to $A_f$, as explained in~\cite{Huang_2022}. 
\\ \textbf{Tasks:}
We designed $4$ toy assembly examples with number of actions varying from $n=8$, to $n=32$. For each experiment, we generated a random $HTM$, composed of $\frac{n}{4}$ joint actions and $\frac{n}{2}$ only robot actions. The remaining actions can be performed by either the human and the robot. The nominal action duration is randomly generated between $4\bar{T}$ and $16\bar{T}$.
Additionally, we considered the simplified but realistic example of the IKEA Ivar chair assembly, whose HTM is reported in Fig~\ref{fig:HTM_real}. 
\\ \textbf{Baselines:} DECAF's performance is benchmarked against that of two baselines policies: 
the first is a greedy policy inspired to \cite{fusaro2021human}, which selects, among the feasible actions, the one with the lowest nominal duration. The second, instead, is a random policy which randomly selects the next robot action from the feasible actions set with uniform probability. 
\subsection{Deterministic setting}
Tab.~\ref{table:det} shows the assembly duration distributions obtained with the DECAF policies, RL and Graph, and the baselines, Greedy and Random, on the simulated tasks described above and with no uncertain events. The variability in task execution is due to the random human behaviors and the stochastic actions duration.
\begin{table}[hb!] 
    \scriptsize
    \centering
    \begin{tabular}{ c c c c c c}
    \hline \\ [-1.5ex]
    Method& 8 Actions & 16 Actions & 24 Actions & 32 Actions  & Chair\\ 
    \hline \\  [-1.5ex]
   RL  & 99.4 [1.2]
      & 175.1 [1.7]
      & 274.0 [2.1]
      & 340.7 [2.3]
      & 134.6 [1.5]
     \\ [0.5ex]
   Graph  & 99.6 [1.3]
      & 174.9 [1.5]
      & 274.6 [2.3]
      & 340.9 [2.3]
      & 135.0 [1.4]
     \\ [0.5ex]
   Greedy  & 103.4 [4.2]
      & 186.0 [4.8]
      & 275.7 [9.3]
      & 357.9 [10]
      & 135.9 [4.3]
     \\ [0.5ex]
    Random  & 117.9 [11]
      & 201.8 [13]
      & 285.8 [13]
      & 372.2 [14]
      & 138.1 [4.5]
     \\ [0.5ex]
     
     \hline
     
    \end{tabular}
    \caption{Completion time on the deterministic experiments. The table reports the mean and the standard deviation (between square brackets) in terms of number of time steps $\bar{T}$, obtained with $1000$ trials.}
    \label{table:det}
\end{table}

In all the $5$ considered tasks, the RL policy produces almost the same results of the Graph policy, which confirms that RL is able to learn the optimal policy. Moreover, the DECAF policies outperforms both the Greedy and the Random planners. Note that the Greedy policy selects the robot action that has the lowest atomic cost without considering the overall assembly cost. In contrast, DECAF policies optimize over all the possible execution paths which leads to improved coordination and better performance. Moreover, the difference in performance is more evident at the increase of the task complexity, where a large number of different assembly paths are possible. As one could expect, the random policy is not optimal and thus leads to in-average worse results, with larger standard deviations.

\subsection{Stochastic setting}\label{sec:stocastic}

In the stochastic setting, we introduce human change of mind and action failures. These events introduce stochasticity in the possible execution paths, which makes impossible or very inefficient to build the decision graph. For this reason, in this scenario we consider only the RL policy. 

Tab.~\ref{table:CoM} shows the execution time at the increase of the Human Change of Mind probability $p_c$, with $p_c$ ranging from $0.1$ to $0.4$. For each value of $p_c$, we trained a PPO policy and tested it with a Monte Carlo experiment composed of $1000$ simulations. The RL policy demonstrates its effectiveness in addressing changes of mind by computing viable solutions in all cases. As expected, at the increase of the change of mind probability, the average completion time increases.

\addtolength{\tabcolsep}{-2pt}  
\begin{table}[ht!] 
    \scriptsize
    \centering
    \begin{tabular}{ c c c c c c}
    \hline \\ [-1.5ex]
    $p_c$& 8 Actions & 16 Actions & 24 Actions & 32 Actions  & Chair\\ 
    \hline \\  [-1.5ex]
   0.1  & 100.2 [3.2]
      & 177.7 [4.4]
      & 277.9 [5.8]
      & 344.8 [6.2]
      & 135.0 [6.0]
     \\ [0.5ex]
   0.2  & 101.5 [4.8]
      & 180.2 [6.0]
      & 279.7 [7.5]
      & 363.1 [6.2]
      & 139.9 [9.5]
     \\ [0.5ex]
   0.3  & 102.7 [6.0]
      & 183.2 [7.6]
      & 281.5 [11.1]
      & 359.5 [15.2]
      & 144.1 [12.9]
     \\ [0.5ex]
    0.4  & 103.9 [6.8]
      & 187.8 [10]
      & 286.5 [15.1]
      & 380.6 [17.5]
      & 151.0 [16.8]
     \\ [0.5ex]
     
     \hline
     
    \end{tabular}
    \caption{Completion time at the increase of change of mind probability. The table reports mean and standard deviation (between square brackets) in terms of number of time steps $\bar{T}$, obtained after $1000$ trials  with the DECAF RL~policy.}
    \label{table:CoM}
\end{table}
\addtolength{\tabcolsep}{2pt}

Similarly, Tab.~\ref{table:failure} shows the execution time at the increase of the action failure probability. For simplicity, we considered each action to have the same probability of failing, $p_f \in [0.1,0.4]$. The occurrence of a failure introduces in the feasible action set a recovery action, which needs to be performed to complete the previously failed action. The DECAF framework with RL policy leads to a feasible assembly plan also with failures and recovery actions. As expected, also in this case the overall average assembly time is increased at the increase of the failure probability.

\addtolength{\tabcolsep}{-2pt}  
\begin{table}[ht!] 
    \scriptsize
    \centering
    \begin{tabular}{ c c c c c c}
    \hline \\ [-1.5ex]
    Failure p& 8 Actions & 16 Actions & 24 Actions & 32 Actions  & Chair\\ 
    \hline \\  [-1.5ex]
   0.1  & 111.3 [15.0]
      & 196.8 [17.7]
      & 327.6 [22.6]
      & 511.2 [26.6]
      & 149.4 [13.6]
     \\ [0.5ex]
   0.2  & 125.4 [19.1]
      & 222.7 [24.9]
      & 400.5 [30.1]
      & 556.7 [34.2]
      & 174.0 [17.9]
     \\ [0.5ex]
   0.3  & 137.6 [21.6]
      & 295.7 [28.5]
      & 453.1 [33.6]
      & 602.9 [37.9]
      & 184.2 [21.4]
     \\ [0.5ex]
    0.4  & 153.0 [23.2]
      & 400.5 [30.1]
      & 485.5 [36.6]
      & 649.7 [41.7]
      & 209.6 [21.1]
     \\ [0.5ex]
     
     \hline
     
    \end{tabular}
    \caption{Completion time at the increase of action failure probability. The table reports mean and standard deviation (between square brackets) in terms of number of time steps $\bar{T}$, obtained with $1000$ trials  and the DECAF RL policy.}
    \label{table:failure}
\end{table}
\addtolength{\tabcolsep}{2pt}

\section{Real User Study}\label{sec:real_experiment}
DECAF efficacy is validated on a real world experiment involving human subjects. 
The experiment consisted in the human participants assembling an IKEA Ivar chair together with a 7 Degrees of freedom Franka Emika Panda robotic arm. As a baseline, participants undertook the assembly task individually to gauge the enhancements in both performance and work experience resulting from the introduction of the robot equipped with the DECAF planner.

\subsection{Experimental Setup and Task Description}
\begin{figure}
    \centering
    \includegraphics[width=.9\columnwidth]{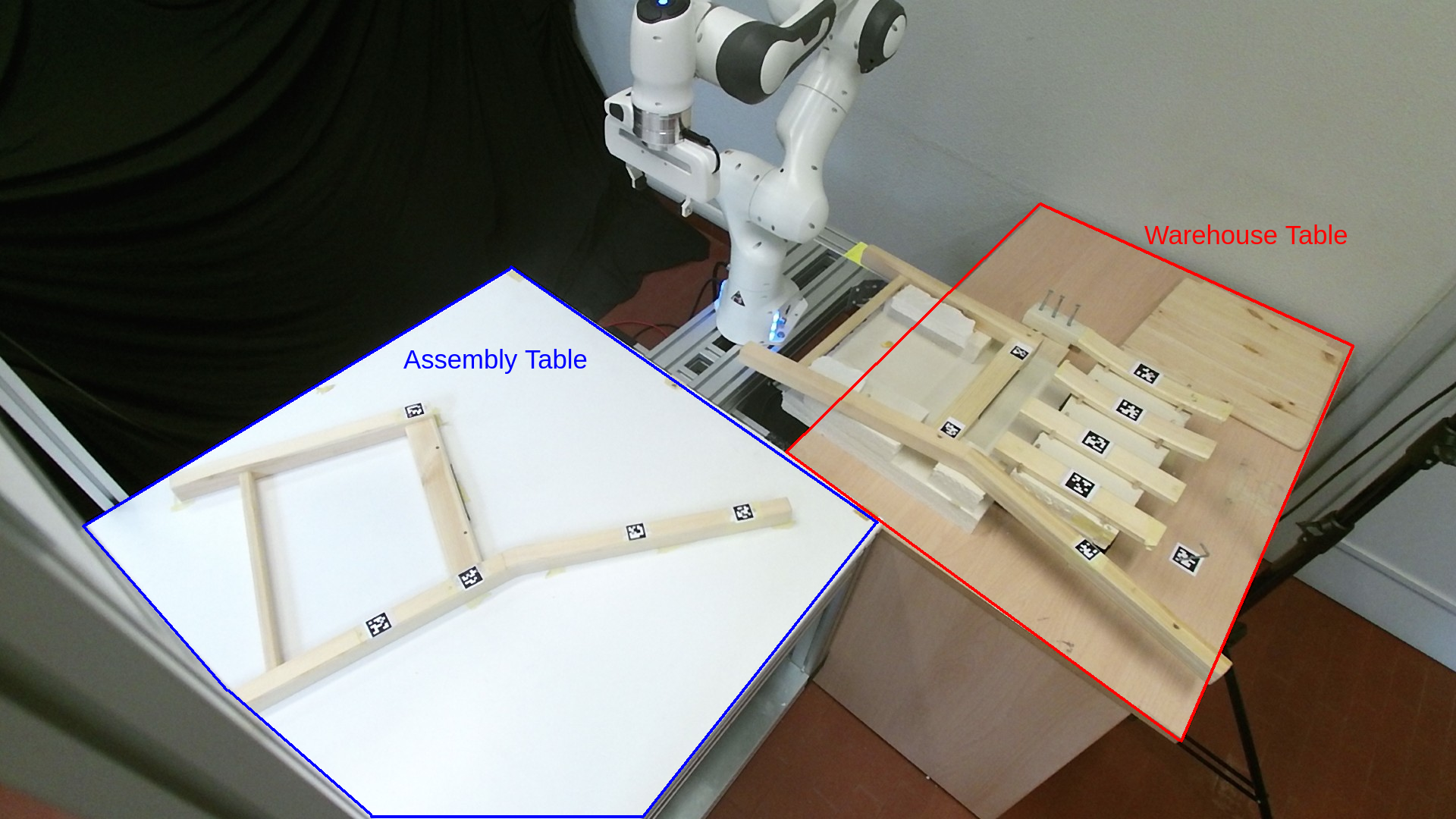}
    \caption{Experimental setup for the Ivar chair assembly task.}
    \label{fig:iaslab_setup}
\end{figure}

The experimental setup consists of the robotic workcell shown in Figure~\ref{fig:iaslab_setup}, including a Franka Emika Panda manipulator and two tables: a \emph{warehouse table} where the components to be assembled are stored, and an \emph{assembly table} where the collaborative assembly process takes place. The area of each table is monitored by an RGB-D camera (i.e., Intel Realsense LiDAR Camera L515), using AprilTag~\cite{Malyuta2019} fiducial markers to detect and localize all the various chair components.  
As shown in Figure~\ref{fig:intro_fig}, the chair is composed of several individual parts: left and right side, a wooden panel for the seat and rails connecting the two sides through the use of dowel pins and screws.
In order to simplify rails insertion operations by the robot, the various dowel pins were replaced by neodymium magnets.

\begin{figure}
    \centering
    \includegraphics[width=.8\columnwidth]{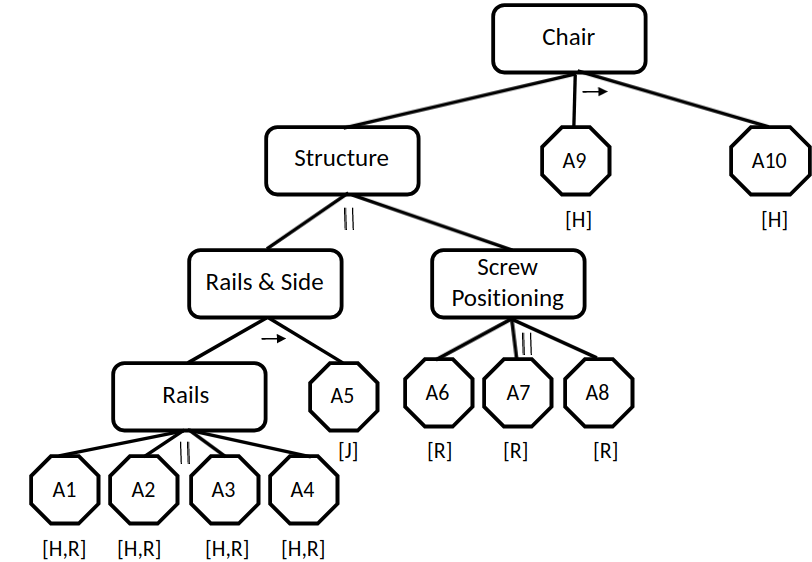}
    \caption{The HTM of the Ivar chair assembly task.}
    \label{fig:HTM_real}
\end{figure}

The assembly process can be summarized as follows: (i) initially the right side of the chair is on the assembly table; (ii) one at a time the rails are placed in their respective seats on the side of the chair; (iii) the left side of the chair is placed on top of the rails; (iv) one at a time the screws are places on top of the left side part; (v) the rails and the side part are screwed together with an Allen key; (vi) the wooden seat is positioned. The HTM representing the chair assembly task is reported in Fig.~\ref{fig:HTM_real}.

The task is composed by the following actions: (i) pick and place of the rails, which can be assigned to either the human or the robot (A1-A4); (ii) collaborative transportation of the left side of the chair, where the robot helps the human carrying the weight and assists in the precise positioning of the side (A5); (iii) pick and place of the screws, a robot-only operation due to the significant distance between the human operator and the screws on the warehouse table (i.e., to promote a better ergonomics). The robot places the screws based on the information provided by AprilTag markers. (A6-A8); (iv) screwing, a human-only operation where the human operator tighten the screws placed by the robots (A9); (v) placement of the seat, a human-only action (A10). 

All the robot actions are implemented as pick and place motion primitives by means of ROS~\footnote{https://www.ros.org/} and MoveIt~\footnote{https://moveit.ros.org/}. The cameras are extrinsically calibrated with respect to the robot base by means of hand-eye calibration~\cite{evangelista2022unified}, allowing to express all the AprilTag poses detected by the cameras in the robot reference frame.

\subsection{Participants and Experiment Design}
The experiments involved 10 volunteers (3 women and 7 men) aged between 24 and 46. The group consists of 3 non-robotic experts and 7 experts in the general field of robotics. Each participant performed 3 trials on the system: one trial assembling the chair alone without any collaboration with the robot, and two trials in collaboration with the robot arm.
Upon executing all trials, users completed two short surveys: the NASA-TLX \cite{hart1988development} to evaluate the workload and a custom Likert scale (5 points) questionnaire (Figure~\ref{fig:rbt_questionnaire}) to evaluate the user satisfaction of the collaborative assembly with robot intervention.

\subsection{Results}
We assess collaboration by examining assembly completion time and the outcomes of the aforementioned questionnaires. Our objective is to determine whether introducing the robot equipped with the DECAF planning framework not only enhances execution time but also improves the quality of the assembly experience from the user's perspective.

\begin{figure}[h!]
    \centering
    \includegraphics[width=.8\columnwidth]{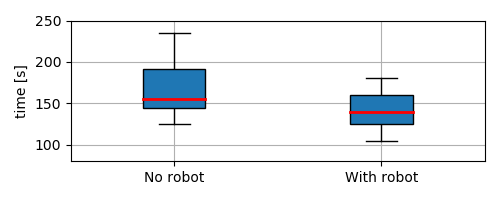}
    \caption{Completion time distribution obtained with the real user experiments described in section \ref{sec:real_experiment}.}
    \label{fig:real_completion_time}
\end{figure}

Fig.~\ref{fig:real_completion_time} compares the completion time distribution obtain by the $10$ subjects when assembly the chair alone or with the robot. It is important to note that integrating a robotic collaborator into the assembly process has introduced certain delays. Understanding the robot's intentions typically requires the user to pause and observe it, while sharing the workspace with a moving robot necessitates slower human movements. Furthermore, since our primary focus is not on implementing optimal robot movements, the robot did not operate at maximum speed during testing. Consequently, the joint actions, such as frame transportation, were performed more slowly by the human-robot team compared to a solo human performer. Nevertheless, the average total assembly time remains lower with the robot than without, confirming the advantages of incorporating the robot from an efficiency standpoint. In summary, although there are minor drawbacks related to collaborating with a robot agent, the net result shows reduced overall assembly times, indicating a positive impact on productivity.

\begin{figure}[h!]
    \centering
    \includegraphics[width=.95\columnwidth]{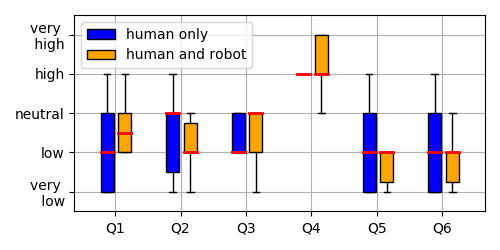}
    \caption{Nasa TLX questionnaire.
            \textbf{Q1}:~How mentally demanding was the task? 
            \textbf{Q2}:~How physically demanding was the task?
            \textbf{Q3}:~How hurried or rush was the pace of the task?
            \textbf{Q4}:~How successfully were you in accomplishing the task?
            \textbf{Q5}:~How hard did you have to work to accomplish the task?
            \textbf{Q6}:~How insecure, discouraged, irritated, stressed, or annoyed were you?
            }
    \label{fig:tlx_questionnaire}
\end{figure}

Fig.~\ref{fig:tlx_questionnaire} summarizes the distribution of the answers to the NASA-TLX questionnaire. Results show that for the participants working with the robot was in general less physically demanding and less hard for the user. However, for some of the participants working with the robot was more mentally demanding than working alone (see answers to Q1 and Q4). As anticipated, we attribute this fact to the problem of sharing the workspace with a moving robot, as well as understanding the robot's intention, which could be counter-intuitive for a non-experienced user.

\begin{figure}[h!]
    \centering
    \includegraphics[width=.95\columnwidth]{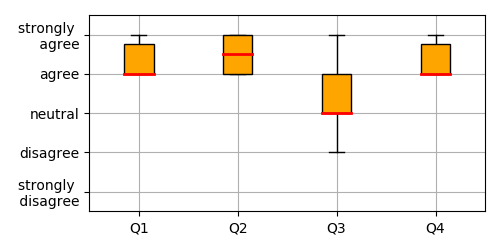}
    \caption{Robot experience questionnaire. 
            \textbf{Q1}:~The robot and I collaborated fluently together to accomplish the task.
            \textbf{Q2}:~I feel the robot had a good understanding of the task.
            \textbf{Q3}:~I was never surprised by the robot actions.
            \textbf{Q4}:~I feel satisfied with the performance of the system.}
    \label{fig:rbt_questionnaire}
\end{figure}

Finally, the results of the custom questionnaire reported in Fig~\ref{fig:rbt_questionnaire} show that the collaboration with the robot has been appreciated by all the participants. Some of them felt surprised by the robot's action choice. We attribute this to the fact that the robot plan is optimized only considering the nominal action duration. This result in a process of action choice that is very different from the decision process of the human, which in turn expect the robot to reason in a human-like fashion. Some user, for example, expected the robot to perform all similar actions consecutively (e.g. position all the screws, then position all the rails), and also would prefer the robot to perform actions uncomfortable for them (e.g. involving objects difficult to reach).
A possible way to mitigate this discordance is the inclusion of metrics measuring for example the ergonomy or the action correlation in the reward function. We leave such investigation as a matter of future work.

\section{Conclusions and future works}\label{sec:conclusion}
In this paper we proposed the DECAF framework for task planning in human robot collaborative assembly. The framework relies on an HTM description of the task alongside a Bayesian inference module for human intent recognition. We have formalized the collaborative assembly between a Human and a Robot as a Discrete Event Markov Decision Problem, which models at the same time all the properties of Parallel-, Sequential-, Joint- Independent-, Asynchronous- and with Variable Duration- actions, and includes Human intent detection change of mind and handling of failures. The formulation as DE-MDP also makes the problem treatable with RL which solves the problem of computational complexity of graph based solutions. We validated the framework in simulations both on toy problems and on a real IKEA Ivar Chair, and we show that the DECAF framework outperforms standard greedy and random planning policies. Moreover, we performed and extended evaluation with real human users, which confirmed the benefits of DECAF both in terms of efficiency and quality of the collaboration. 

In future we plan to include other optimal metrics beyond the execution time, such as human safetiness, action correlation and human ergonomics.

\bibliographystyle{IEEEtran}
\bibliography{references}
\end{document}